\documentclass[runningheads]{llncs}

\usepackage[T1]{fontenc}
\usepackage{graphicx}
\usepackage{fancyhdr}
\usepackage{url}
\usepackage{amsmath, amssymb} 
\usepackage{geometry} 

\fancypagestyle{firstpageheader}{
    \fancyhf{} 
    \fancyhead[C]{\footnotesize PREPRINT: This is a preprint of the following paper, accepted by the 13th International Conference on Model-Based Software and Systems - MODELSWARD25.}
}

\begin{document}

\title{HyperGraphOS: A Meta Operating System for Science and Engineering}

\author{Antonello Ceravola\inst{1}\orcidID{0000-0002-1075-459X} \and
Frank Joublin\inst{1}\orcidID{0000-0002-4421-1737} \and
Ahmed R. Sadik\inst{1}\orcidID{0000-0001-8291-2211} \and
Bram Bolder\inst{1}\orcidID{0009-0002-5595-2466} \and
Juha-Pekka Tolvanen\inst{2}\orcidID{0000-0002-6409-5972}}

\authorrunning{Ceravola et al.}

\institute{Honda Research Institute Europe, Offenbach, Germany \\
\email{\{antonello.ceravola, frank.joublin, ahmed.sadik, bram.bolder\}@honda-ri.de} \and
MetaCase, Jyväskylä, Finland \\
\email{jpt@metacase.com}}

\maketitle

\thispagestyle{firstpageheader}

\begin{abstract}
This paper presents HyperGraphOS, an innovative Operating System (OS) designed for the scientific and engineering domains. It combines model-based engineering, graph modeling, data containers, and computational tools, offering users a dynamic workspace for creating and managing complex models represented as customizable graphs. Using a web-based architecture, HyperGraphOS requires only a modern browser to organize knowledge, documents, and content into interconnected models. Domain-Specific Languages (DSLs) drive workspace navigation, code generation, AI integration, and process organization.

The platform’s models function as both visual drawings and data structures, enabling dynamic modifications and inspection, both interactively and programmatically. HyperGraphOS was evaluated across various domains, including virtual avatars, robotic task planning using Large Language Models (LLMs), and meta-modeling for feature-based code development. Results show significant improvements in flexibility, data management, computation, and document handling.

\end{abstract}

\section{\uppercase{Introduction}}
\label{sec:introduction}

\begin{figure}[t]
  \centering
  \includegraphics[width=\textwidth]{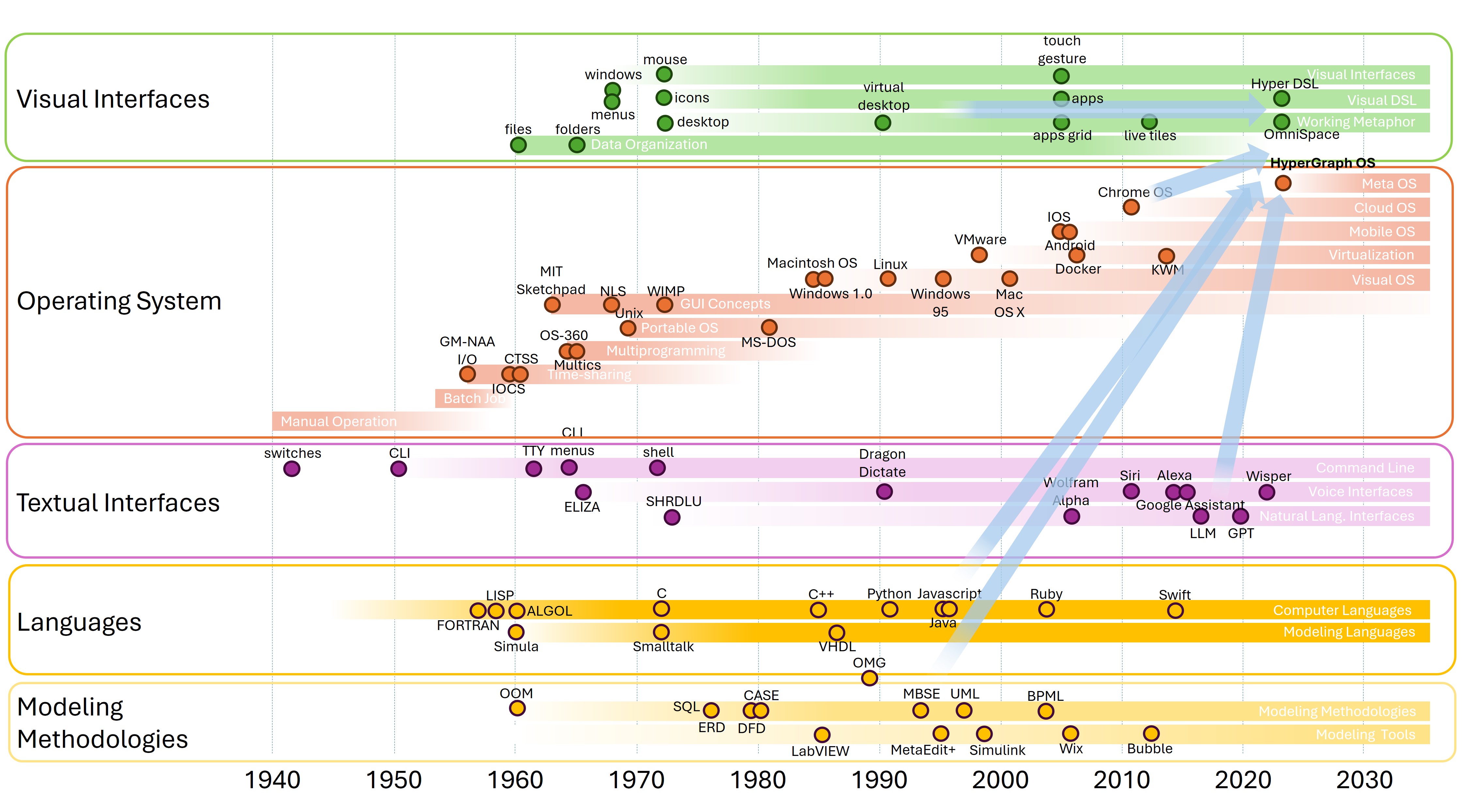}
  \caption{\small Historical Influences leading to the development of HyperGraphOS.}
  \label{fig:hgInfluences}
\end{figure}

Operating Systems (OSs) have evolved significantly since the 1950s, when they were first developed for general-purpose computers such as IBM's 701 and 709, as illustrated in Figure~\ref{fig:hgInfluences}. Initially, these systems required manual intervention for executing programs and lacked automation. The introduction of \textit{batch processing} in the 1950s, exemplified by IBM systems, allowed sequences of jobs to be processed without human input. \textit{Time-sharing} systems soon followed, enabling multiple users to interact with the same computer simultaneously, as seen in MIT’s CTSS and IBM’s System/360. At this same time, Teletypewriters (TTY) and the concept of \textit{file} were introduced, followed a few years later by the concept of hierarchical \textit{folders} in Multics. At the end of the 1960s, UNIX, developed at Bell Labs, popularized these abstractions and the concept of \textit{console} which since then form the foundation of modern OSs.

\section{\uppercase{Background}}
\label{sec:background}

OSs can be defined from several perspectives. However, their primary function is to manage and allocate hardware resources such as memory, processors, and input/output devices, to ensure efficient interactions between users and applications \cite{tanenbaum2009modern,silberschatz2013operating}. An OS abstracts the underlying hardware and resources, providing a user-friendly interface and offering basic services that facilitate interactions for both users and programs.

OSs are generally divided into two categories: general-purpose and special-purpose OSs \cite{bullynck2018operating}. This paper focuses on general-purpose OSs like Windows, Linux, and macOS, designed for broad use and enabling users to organize documents and applications without requiring specific technical expertise. These systems are widely employed in various environments, from households to professional and technical settings, where they support tasks such as document management, billing, and software development. However, general-purpose OSs often fall short when addressing specific domain needs. General-purpose OSs rely exclusively on applications to solve the domain-specific needs of users. For example, household users may find the file and folder structure cumbersome, while professionals may struggle to relate documents like bills and orders without relying on billing applications. Technical users may face challenges in organizing their work environments due to limited project-specific support in traditional OSs if they do not turn to project management applications. The challenge here is that these applications are vendor-dependent and create a zoo of problems (e.g., file formats, compatibility, interoperability) when integration of multiple tools is needed. These problems can all be solved through the use of "glue" applications (e.g., converters) that increase usage complexity in unnecessary ways (accidental complexity \cite{brooks1987no}).

From Model-Based System Engineering (MBSE) perspective \cite{david2023blended}\cite{tolvanen2016model}, OSs can be seen as applications that provide specific DSLs for users to model their tasks and interactions. In MBSE, abstract models represent system architecture, behavior, and interactions without focusing on implementation details \cite{tolvanen2016model}. OSs can be analyzed through their DSLs, which facilitate user interaction, program execution, and hardware management. High-level DSLs define elements like files, folders, and windows, enabling visual organization and interaction with the system. Files and folders, represented on the desktop, have attributes like names, sizes, and creation dates, while windows display application content and include attributes such as size, position, and title. This abstraction simplifies user interaction by hiding the underlying complexity. However, traditional DSLs come with several limitations. \textit{Desktops} introduced in the 1970s followed a working environment metaphor and were extended in the 1990s by the concept of \textit{virtual desktops}. Although they provide space for organizing files, they are restricted by the physical screen size, and the icons often lack sufficient visual clarity for efficient navigation. Furthermore, the desktop layout does not persist after a system reboot, requiring users to manually restore their application layouts—a problem recently mitigated in Windows through tools like PowerToy App Layout \cite{microsoft_powertoys}. While files and folders are effective for document organization, they often create inconsistencies. This is due to reliance on user-defined naming conventions and the lack of flexibility in managing file relationships. Consequently, handling large volumes of files becomes challenging without advanced organizational tools.

The application-centric nature of traditional OSs presents several challenges. Data reusability across different applications often requires tedious or complex conversions, and frequent context switching between applications complicates workflows. Resource management is also handled independently by applications, often resulting in redundant or inefficient use of data storage. Moreover, the heavy reliance on GUIs limits automation and integration with advanced systems, as many embedded high-level DSLs, such as the Automator tool for macOS \cite{apple_automator} and Atlassian Workflow Automation \cite{atlassian_workflow_automation}, are not designed for low-code programming, restricting the ability to automate tasks or integrate with external systems. In Agile Modeling terms, interacting with an OS through its graphical interface is analogous to modeling. Users create "models" of their desired content and actions through the UI, which the OS interprets and executes, updating the system state accordingly. DSLs abstract underlying complexity, enabling users to focus on tasks without dealing with low-level system management. For example, low-level DSLs, such as peripheral APIs, allow OEMs to create new devices for computers. Mid-level DSLs, such as programming APIs, enable developers to build applications without directly managing hardware. High-level DSLs, such as files and folders, allow users to intuitively organize their work based on real-world metaphors like the desktop, modeled after a physical desk. By examining operating systems through an MBSE lens, their limitations, and how DSLs can enhance user interaction, application management, and overall system efficiency, can be better understood.

\section{\uppercase{HyperGraphOS Concept}}

\begin{figure*}[t]
  \centering
  \includegraphics[width=12cm]{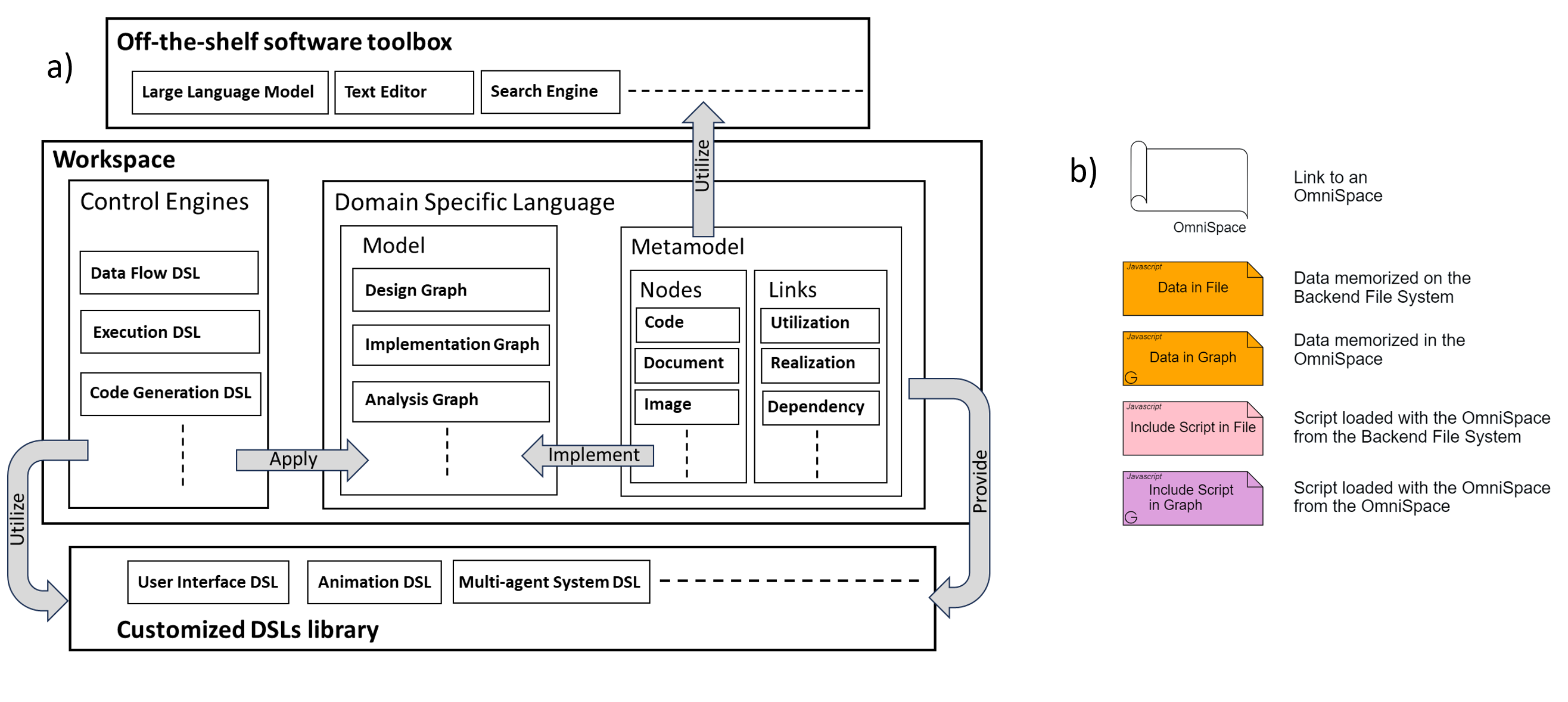}
  \caption{\small a) HyperGraphOS operation concept. b) Basic DSL for navigation and file manipulation}
  \label{fig:hgConcepts}
\end{figure*}

HyperGraphOS \cite{hypergraphos-documentation} is designed to redefine how users interact with computers and digital information systems by leveraging DSLs \cite{hypergraphos-documentation}.  Particularly for interdisciplinary systems like Multi-Agent Systems (MAS). Its core architecture leverages flexible model execution strategies, allowing for rapid prototyping and iterative design adjustments. The tool includes features such as automated code generation, support for multiple modeling paradigms, and a user-friendly interface. The complete implementation of HyperGraphOS, including source code, user guides, and examples, can be accessed through its open-source repository at \cite{hypergraphos-repo}.

The operation concept of HyperGraphOS shown in Fig.~\ref{fig:hgConcepts}a transforms traditional file management into an interconnected web of information. Nodes within the system represent files, documents, and data, which are customizable, annotatable, and maintainable. These nodes visually link data while encapsulating both content and visual aspects, in a concept similar to Unix symbolic links. The semantics of nodes can represent abstractions such as programs, numbers, images, and text, components or any domain specific concepts while links define relationships like dependencies, causal connections, interactions or any domain specific functionalities \cite{hypergraphos-documentation}.

The Meta-model allows for the creation of nodes (such as code, documents, and images) and links (such as utilization, realization, and dependency) for users to organize and visualize their data in flexible and dynamic ways. This architecture promotes seamless interaction with various data formats, which are handled by corresponding editors and viewers. At the heart of HyperGraphOS is the concept of OmniSpace, an infinite workspace, which can be configured with different DSLs, for instance, with Data Flow DSL, Execution DSL, or Code Generation DSL. These DSL/engine can be used by a developer to model an application which can be executed in-place, on a batch or deployed to a target computer. Execution DSL utilizes the Customized DSLs Library, which includes specialized DSLs such as the Basic DSL (Fig.~\ref{fig:hgConcepts}b) for navigation and file manipulation, User Interface DSL, Animation DSL, or MAS DSL, providing a rich set of meta-models to developers for their modeling processes.

The DSL component within HyperGraphOS provides a framework for the development of meta-models, which can be used in different forms, like Design Graphs, Implementation Graphs, and Analysis Graphs. These models use the nodes and links provided by the meta-model, allowing the system to tailor the representation of information to the specific needs of the users. HyperGraphOS offers the capability to define DSLs using Meta-DSLs (notice the recursion here), which are implemented within OmniSpaces for creating and applying domain-specific models (a meta-meta model for the creation of DSL is also available to users). 

\begin{figure}[h]
    \centering
    \includegraphics[width=\linewidth]{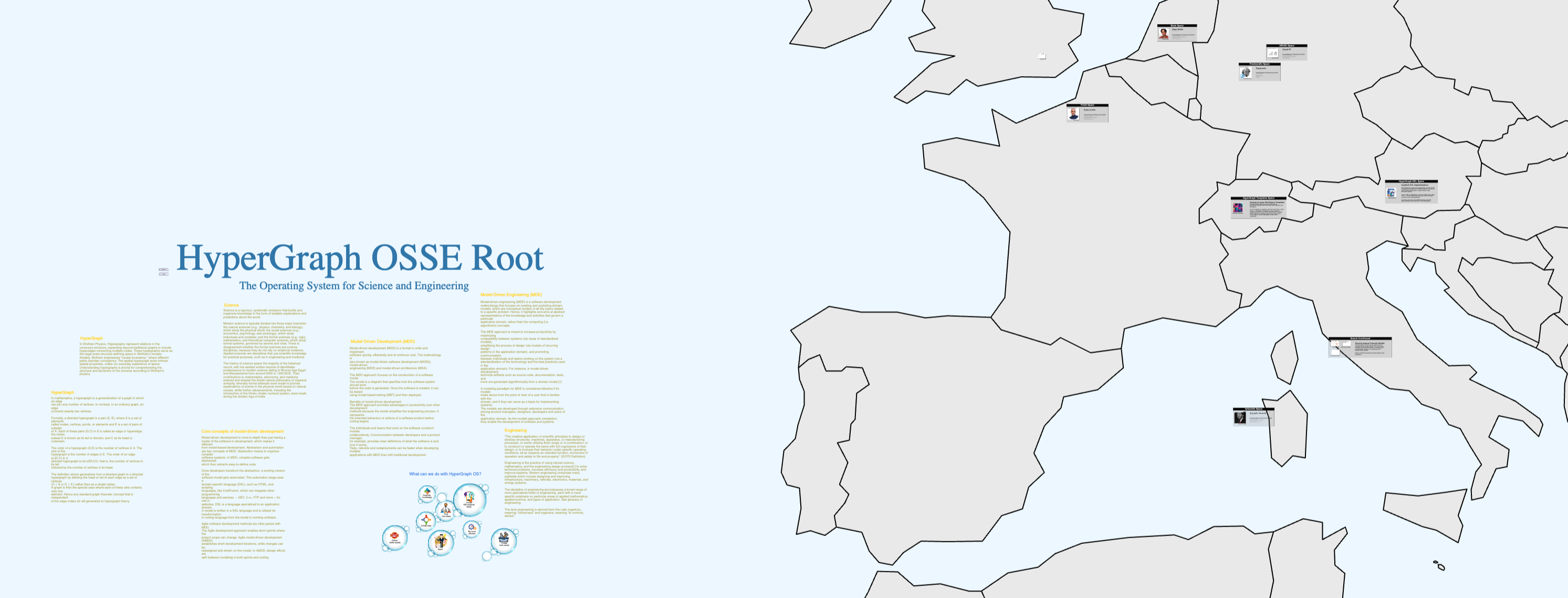}
    \caption{HyperGraphOS Landing Workspace (https://youtu.be/xjoj-snEV\_o)}
   \label{fig:Root-WS}
\end{figure}

HyperGraphOS operates as distributed OmniSpaces (see Fig.~\ref{fig:Root-WS} for the currently used starting OmniSpace), where users can group documents visually using container nodes, links, and OmniSpaces links. OmniSpaces serve as virtual environments for organizing data and applications. They are infinite, flexible, and capable of preserving the state (both model and applications/windows) and navigation history of tasks. OmniSpaces can also span across multiple storage solutions such as local storage, cloud services, and remote devices, allowing for diverse representations and applications. Off-the-shelf Software Toolbox, which includes tools such as LLMs, Text Editors, and Search Engines, are seamlessly integrated into HyperGraphOS. These tools are utilized by the workspace's Control Engines and Meta-model to provide advanced capabilities, such as content creation, search functionalities, and programmatic manipulation.

HyperGraphOS incorporates cutting-edge technologies like an integrated JavaScript-based shell for testing and manipulating nodes and links programmatically. Additionally, a robust search engine and AI integration provide on-demand assistance within documents. The AI assistant, skilled in MBSE \cite{tolvanen2023evaluating}, enhances content creation and modeling processes by suggesting improvements, reading, and writing to models. Despite its rich set of capabilities, HyperGraphOS maintains a minimalistic design and system footprint, ensuring intuitive interaction and ease of use. The concept of applications, in HyperGraphOS, is re-imagined as modular constructs, moving away from the monolithic executable model of traditional OSs. The set of Featured-Based Code Generation Systems allows for automatic code generation based on models, templates \cite{hypergraphos-documentation}, annotations, ... something that further streamlines software development processes.

HyperGraphOS supports AMDD by introducing AI-powered modeling that enhances productivity \cite{sadik2024coding}. Users benefit from AI-generated suggestions and improvements during the code generation phase. Additionally, a data flow DSL and execution engine allow for real-time model execution and facilitate the code generation process. Collaboration is key in HyperGraphOS. Although this is still under-developed, functionalities to enable multi-disciplinary teams to work together using integrated tools for real-time editing, version control, and interactive annotations have been started. This collaboration functions would help fostering synergy among team members from various domains. HyperGraphOS represents a paradigm shift in how users interact with digital systems. Its powerful, flexible, and intuitive platform supports model-based system engineering and development, assisting in overcoming limitations of traditional OSs.

\section{\uppercase{System Architecture}}

\begin{figure*}[t]
  \centering
  \includegraphics[width=11cm]{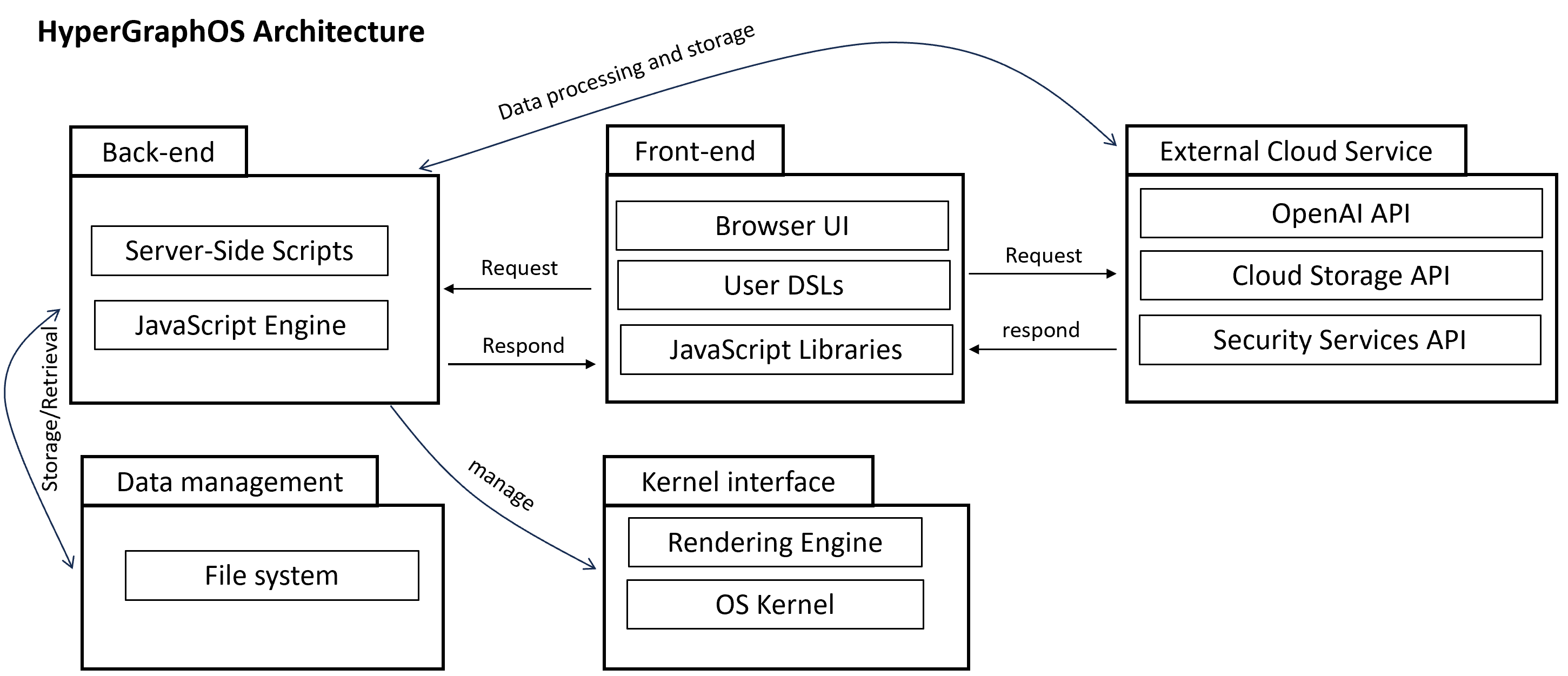}
  \caption{\small HyperGraphOS software architecture}
  \label{fig:hgArchi}
\end{figure*}

HyperGraphOS is built on a modular architecture as shown in Fig.~\ref{fig:hgArchi}. The architecture is composed of five main modules: a Kernel Interface, a Back-end, Front-ends, External Cloud Services, and Data management. These modules work together to provide a flexible, distributed, and scalable system that redefines traditional file management and work organization. At the core of HyperGraphOS is the Kernel Interface, which manages the hardware abstraction and rendering of the user interface. The Rendering Engine within this module ensures seamless graphical interaction, while the OS Kernel interfaces with essential hardware resources like CPU, memory, and input/output devices. Additionally, the File System plays a crucial role in managing data storage and retrieval, interfacing with the back-end for efficient data processing and handling of user OmniSpaces.

The Back-end acts as an intermediary between the front-end and the kernel, built on JavaScript Engine components and Server-Side Scripts that handle data requests, OmniSpaces management, and batch execution. This module processes user requests, manages OmniSpaces files as JSON objects, and ensures that data is stored or retrieved from the file system. Moreover, the back-end is responsible for interacting with External Cloud Services, enabling integration with local or cloud-based APIs like OpenAI for AI tasks, Cloud Storage for scalable data handling, and Security Services for managing data protection and privacy. The Front-end of HyperGraphOS operates through a browser interface, where users interact with OmniSpaces via a graphical canvas powered by DSLs, JavaScript Libraries and GoJS (one of the most complete graphical libraries available in the web domain \cite{gojs2024}). This dynamic interface allows users to manipulate visually the content of OmniSpaces, using graphs to represent nodes, and links. Each workspace is stored as a JSON object, allowing light and flexible storage \cite{nurseitov2009comparison}\cite{zunke2014json} and intuitive management of files and documents (due to easy access of its content through a visual inspection or programmatic one). The front-end ensures that users have a streamlined and interactive experience, directly connected to the back-end for data requests and processing.

The External Cloud Services module integrates key functionalities that extend the capabilities of HyperGraphOS. Through cloud-based APIs, the system interacts with external tools for data processing, security management, and AI-driven features. Services like OpenAI API provide powerful machine learning and natural language processing capabilities, while Cloud Storage offers scalable and distributed storage solutions, ensuring the system can handle an increasing number of OmniSpaces as user needs grow. The Security Services API guarantees user data protection, ensuring privacy and compliance with security standards while maintaining a lightweight system architecture. Data management in HyperGraphOS is streamlined through a custom organization of files and directories on the server side, bypassing for now the need for traditional databases. This approach simplifies data architecture while ensuring flexibility in managing OmniSpaces. However, as HyperGraphOS evolves, further enhancements may be required to accommodate more complex data management needs.

Scalability is inherently managed through the distributed handling of JSON files that represent OmniSpaces. This ensures that the system can efficiently manage multiple nodes or OmnisSpaces without the need for extensive infrastructure, enabling HyperGraphOS to scale seamlessly as users create and manage more complex environments. Security and privacy are handled through external services, allowing HyperGraphOS to maintain a lightweight architecture without sacrificing user data protection. By outsourcing security management to specialized services, the system remains streamlined and efficient, ensuring robust data protection without adding unnecessary complexity to the core architecture.

HyperGraphOS also provides robust integration capabilities through its own APIs, allowing programmatic navigation and modification of WorkSpace models, with models represented in a dual way as a visual drawing and as a JSON object. JSON format has been chosen for its native integration in Javascript and its relative lightweight encoding \cite{nurseitov2009comparison}\cite{zunke2014json}. Besides JSON, some dedicated user interface and DSLs make use of the YAML format \cite{eriksson2011comparison} for its compactness, readability and user friendliness. Both the front-end and back-end offer libraries that support users in defining code generators for their applications, making integration with external tools and systems straightforward and seamless.

This modular and distributed architecture, combined with the flexibility offered by JSON-based workspace management, allows HyperGraphOS to deliver a scalable, secure, and efficient solution for modern computing needs. It offers a new approach to manage digital information by extending the limited concept of file and folders with DSL elements that now can represent a file, a portion of it or a group of them, connected with visible links. In this context new approaches like versioning, will soon be disclosed in further publications, blending the simplicity of user-friendly design with the power of advanced, customizable back-end services.

\section{\uppercase{Case Studies}}

In this section, three case studies are presented to demonstrate the practical application of various system modeling and artificial intelligence methodologies using HyperGraphOS. Each case study highlights the significant contribution that HyperGraphOS provides in the creation of a multi-agent robotic task execution system, a meta-model for system architectures in research applications, and a dialog management system. The first two will be briefly presented, and the last one will be explored in greater detail.

\vspace{0.3cm}
\textbf{Case Study 1: Multi-Agent Robotic Task Planning and Execution}

In this case study, HyperGraphOS is used to develop a robotic control system based on multi-agent task planning and execution \cite{joublin2024copal}. The system integrates natural language processing with task and motion planning using a hierarchical architecture built with OpenAI's LLMs. The CoPAL (Cognitive Planning and Learning) system allows the robot to perform complex tasks in the real world, such as preparing pizza and stacking cubes. This is achieved by incorporating replanning feedback loops. The model, defined using the dataflow DSL, integrates components allowing ROS \cite{quigley2009ros} communication with the robotic system. The research and development of CoPAL with HyperGraphOS demonstrates the flexibility of modeling, executing, debugging, and testing complex data flow models that generate tasks for a humanoid robot in the real world. The core development of the multi-agent system, including the DSL, took only a single week, allowing most of the time to be spent on evaluation and experiments with the model.

\vspace{0.3cm}
\textbf{Case Study 2: Modeling Research Projects with Thebes DSL}

This case study addresses the challenge of managing dynamic research projects using Thebes, a lightweight DSL tailored for modeling flexible research projects within HyperGraphOS. Thebes facilitates rapid prototyping and incremental design, enabling seamless integration with existing tools via code generation. Applied to projects like the tabletop robot Haru \cite{gomez2018haru} and the CoPAL system \cite{joublin2024copal}, Thebes significantly improved collaboration and adaptability. HyperGraphOS provided support for creating the metamodel in about 30 minutes and implementing the model integrity checkers and code generation in JavaScript in about three days.

\vspace{0.3cm}
\textbf{Case Study 3: Virtual Receptionist for Visitor Registration}

This case study focuses on the development of a virtual receptionist system used for visitor registration at a research institute \cite{joublin2024introducing}. The system, initially developed before the widespread adoption of LLMs, utilized a recursive neural network to define a behavior engine for the AI-driven receptionist. This research explored the challenges of creating dialogue systems capable of interacting with users through natural language or speech. Traditional dialogue systems often face several issues, including the need for extensive training data and difficulties in defining reward functions. They also struggle with limited control and explainability.

\begin{figure}
    \centering
    \includegraphics[width=1\linewidth]{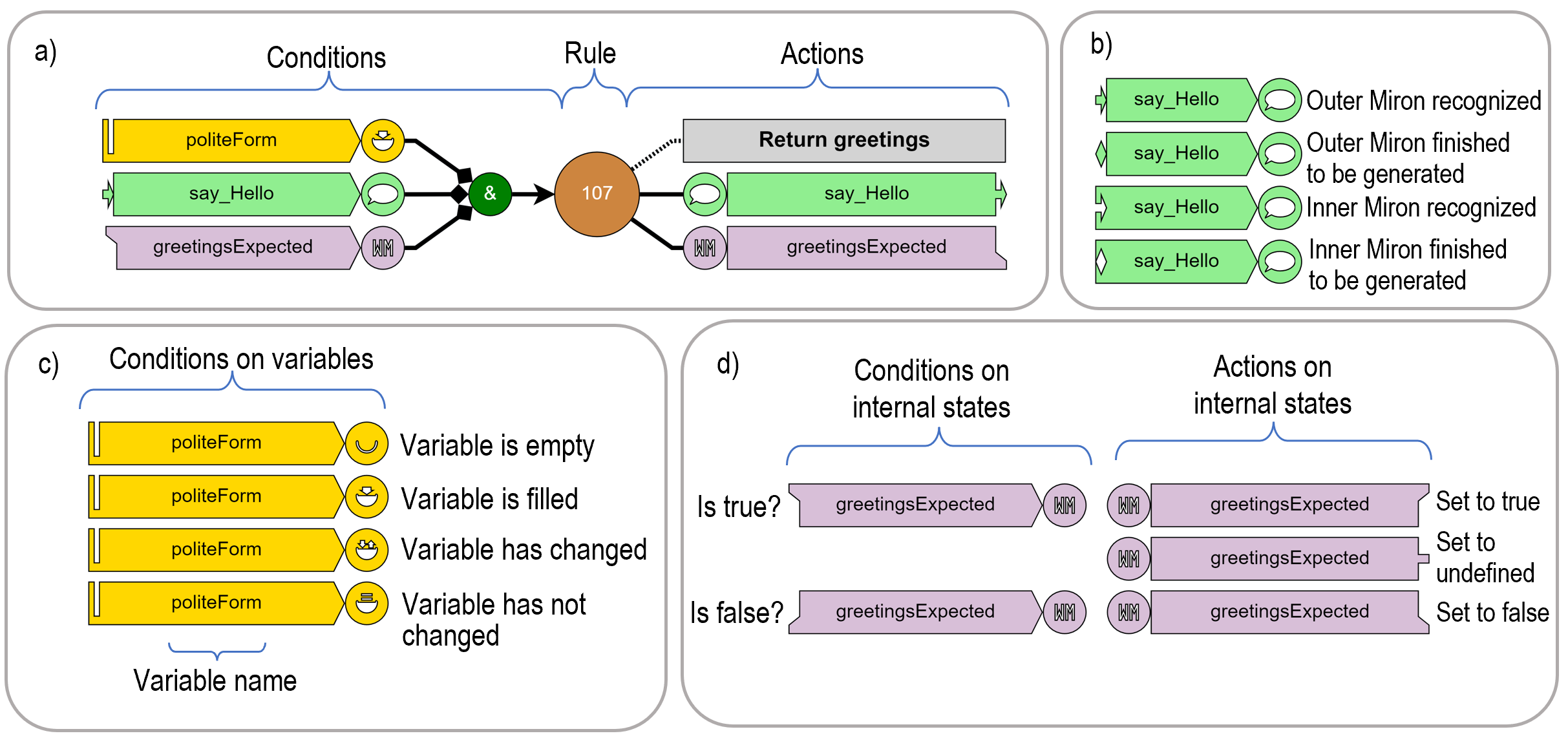}
    \caption{\small Rule DSL Elements : A) Graphical representation of a rule defined by its auto-generated ID (107), its conditions and its actions. The gray box is just a comment used to explain the
rule. B) Possible state of a Miron (an abstraction used to equally recognize and generate sentences) that can be used as part of a condition. C) Possible state of a variable
that can be used as part of a condition. D) Possible conditions and actions on internal states. Link visual aspect is automatically determined by the connected nodes.}
    \label{fig:ruleDSL}
\end{figure}

\begin{figure}
    \centering
    \includegraphics[width=1\linewidth]{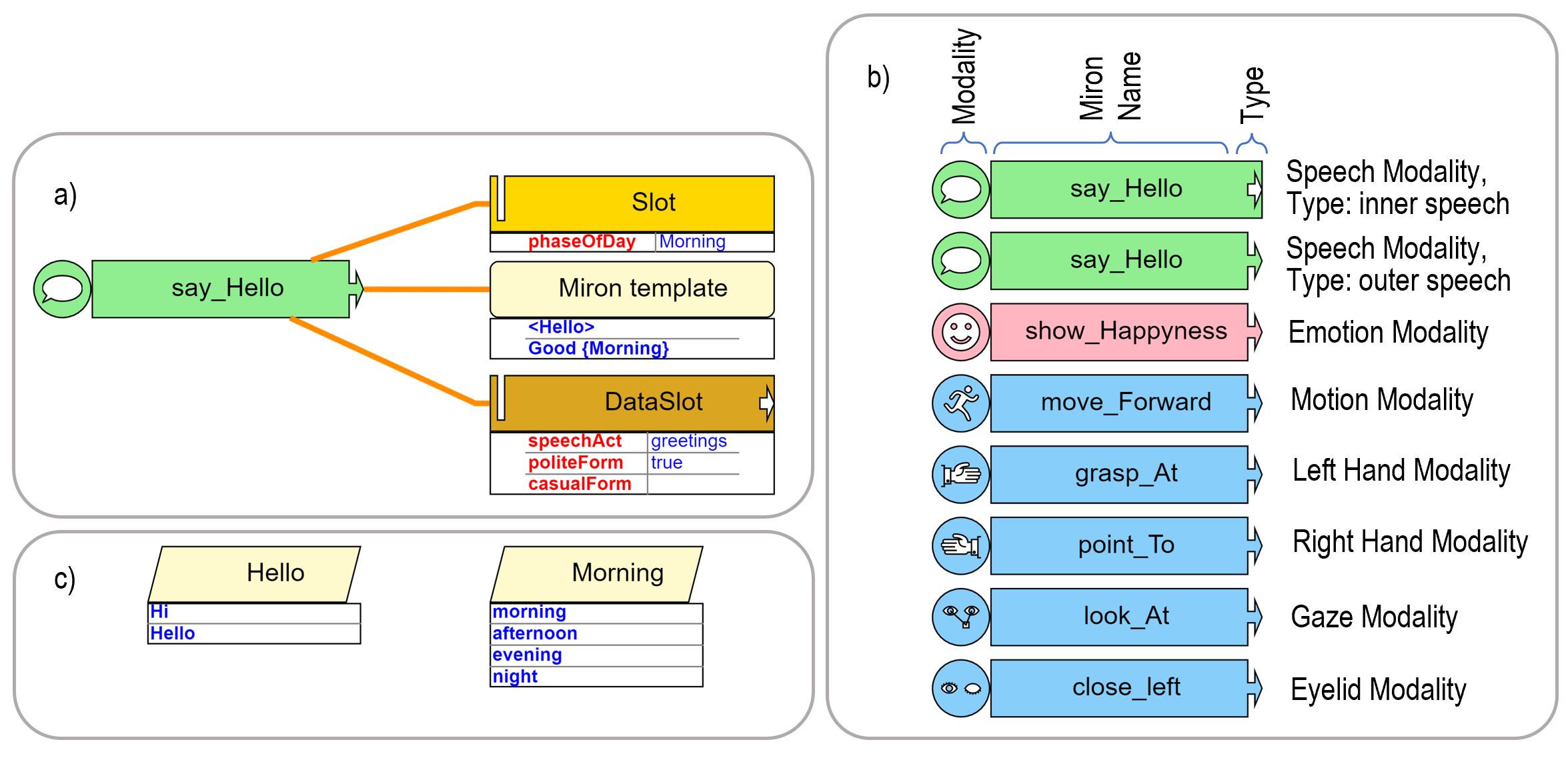}
    \caption{\small Miron DSL elements: A) Graphical representation of a Miron defined by a modality, a name, a type (inner or outer), templates, named entities (slots) and associated data
(data slots). B) Example of different Miron modalities. Modalities were used to control speech output and motion and expression of a virtual avatar. C) Grammar fields defining alternative verbal expressions.}
    \label{fig:mironDSL}
\end{figure}

To address these challenges, the team designed a neural behavior engine inspired by neurobiology and neuropsychology, which incorporated concepts such as mirror neurons and multi-modal embodiment. This engine facilitated mixed-initiative dialog and action generation. The system was successfully implemented as a virtual receptionist in a semi-public space, demonstrating its capability to manage real-world interactions with users.

\begin{figure}
    \centering
    \includegraphics[width=1\linewidth]{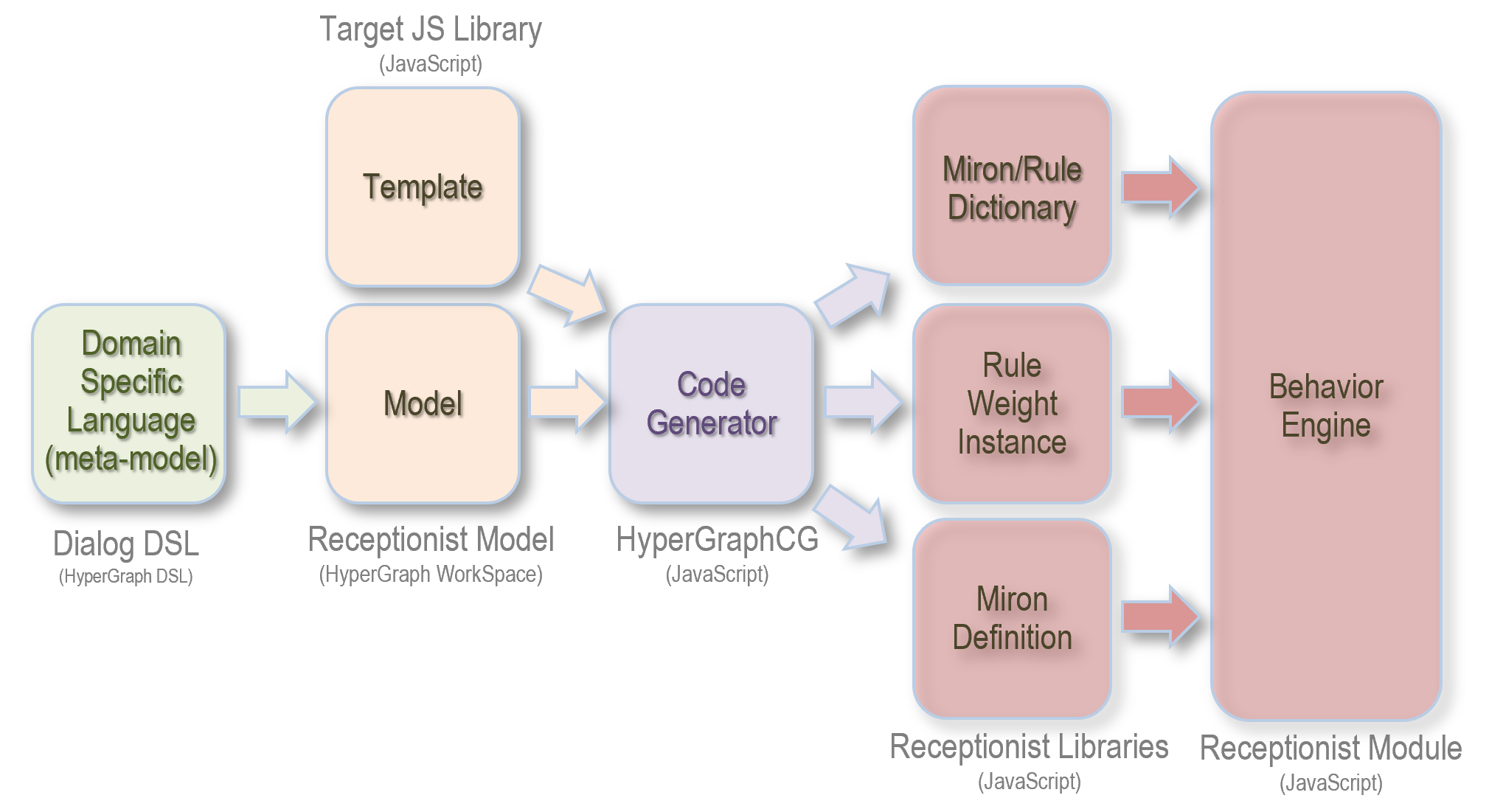}
    \caption{\small Avatar Receptionist Network Modeling Process}
    \label{fig:CGRules}
\end{figure}

HyperGraphOS played a pivotal role in two key aspects of the system's development:

\begin{itemize}
    \item HyperGraphOS was used to define a DSL for the behavior engine together with a code generator. The DSL was created in 3 days and the code generator in one week. The DSL is based on a clock-based architecture model created in a dedicated workspace, and the generated code integrated into a target JavaScript module for the avatar receptionist system.
    \item HyperGraphOS was also employed to design a Dialog DSL (Fig.~\ref{fig:ruleDSL} and ~\ref{fig:mironDSL}) based on parallel state flow. The DSL and the code generation (Fig.~\ref{fig:CGRules}) has been created in about two weeks. The model, implemented in a workspace, consisted of 4246 nodes and 3890 links, which generate JavaScript files such as dictionaries (4033 generated lines), weights for the recurrent network (4659 generated lines), and NLP intents (2410 generated lines). Average code generation time (from reading the model to generating all files) was less than 3 seconds on a 12th Gen Intel Core i7-12700H laptop.

\end{itemize}

\begin{figure}
    \centering
    \includegraphics[width=1\linewidth]{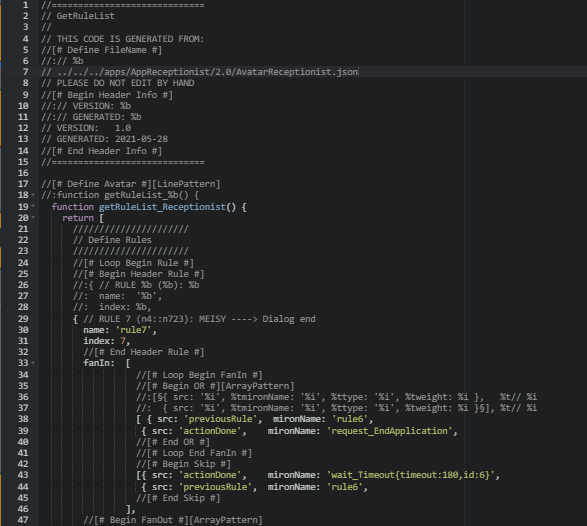}
    \caption{\small Example of Avatar Dialog Network source template. Comments like '//[\# command \#] represent code generation commands, while comments like '//: ...' represent command parameters.}
    \label{fig:cgTemplate}
\end{figure}

To illustrate the development of a DSL and the process of code generation in HyperGraphOS, this case study focuses on the creation of the Dialog DSL (Fig.~\ref{fig:fullModel}). This case study followed the complete DSL creation process in HyperGraphOS to define a specific DSL. The first step involved defining a meta-meta-model to determine the visual representation of each DSL element. HyperGraphOS supports this phase with JavaScript functions leveraging the GoJS model concept. Once the meta-meta-model was established, the meta-model for the Dialog DSL was defined, which can be visually designed within an OmniSpace using HyperGraphOS's dedicated DSL for building DSLs. During this stage, the elements of the Dialog DSL were specified along with their attributes and semantics (see Fig.~\ref{fig:ruleDSL} and ~\ref{fig:mironDSL}). HyperGraphOS automatically adds the user-defined DSL to a system palette for easy access.

Currently, HyperGraphOS offers two primary approaches for defining DSLs: 1) a full process that involves creating both the meta-meta-model (using JavaScript and GoJS) and the meta-model (drawn in OmniSpace), and 2) a light process, where users define a meta-model by parameterizing a DSL creation tool within OmniSpace. Additional methods for defining DSLs are under exploration and will be addressed in future publications.

\begin{figure}
    \centering
    \includegraphics[width=1\linewidth]{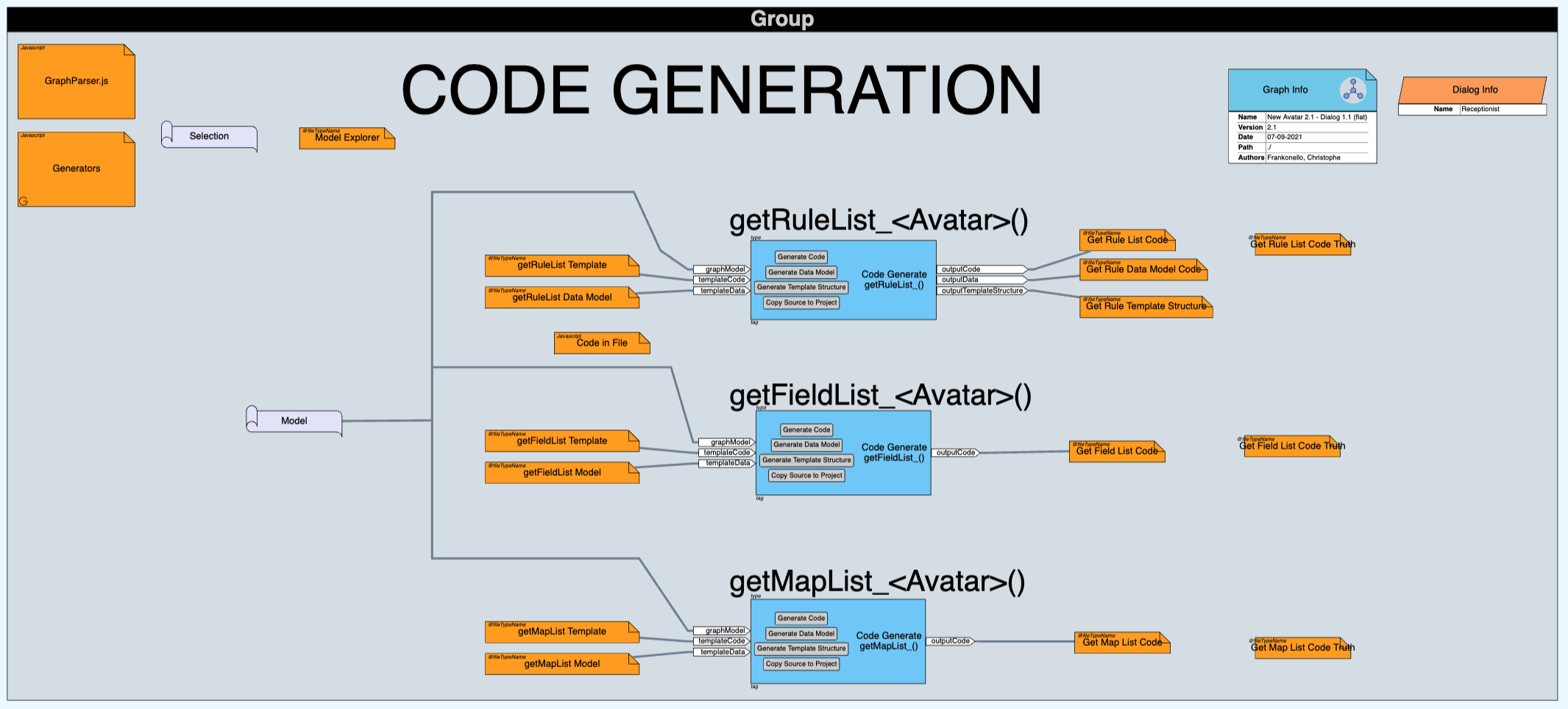}  
    \caption{\small Code generation model for the Dialog Model}
    \label{fig:codeGen}
\end{figure}

\begin{figure*}
    \centering
    \includegraphics[width=1\linewidth]{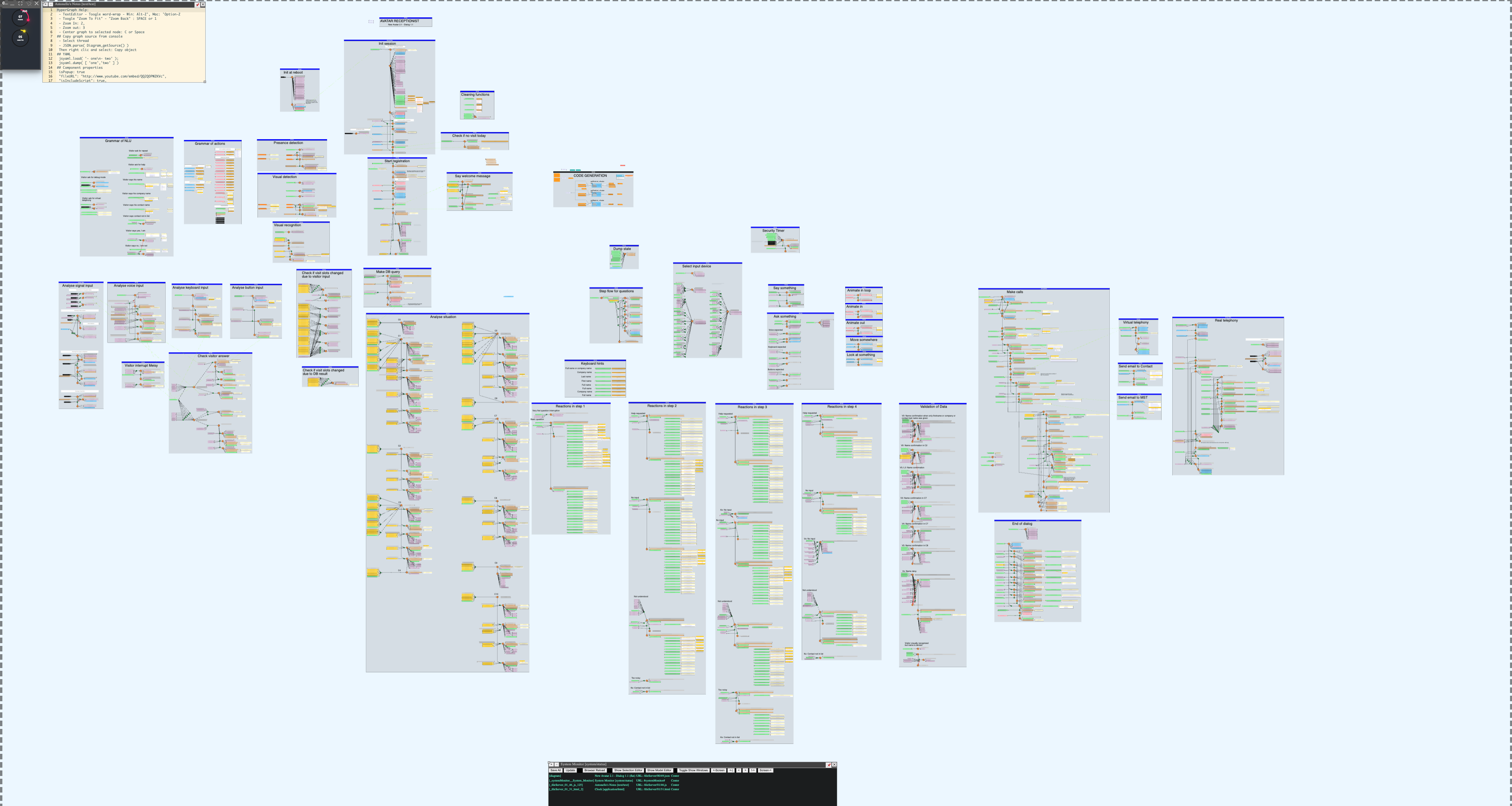}
    \caption{\small View of the full Dialog Model for the Avatar Receptionist. The code generation model shown in figure \ref{fig:codeGen} is in the center top of this model}
    \label{fig:fullModel}
\end{figure*}

For code generation in the Dialog Model created for this application, one of HyperGraphOS's template engines was used. HyperGraphOS provides several template generators, which can be extended by users. The process involves the following steps: first, a target example file is required, serving as a running example of the code to be generated. The example file is then transformed into a template by adding annotations as comments (see Fig.\ref{fig:cgTemplate}). Next, the code generation logic is defined in a model (see Fig.\ref{fig:codeGen}), where it is implemented.

The system demonstrated robustness in real-time scenarios, effectively managing dialog states and context, and seamlessly switching between different modalities (e.g., speech or text interaction, or a combination of both, including telephony) while gracefully handling errors. The use of DSLs for behavior and engine modeling facilitated scalability and maintainability, ensuring ease of maintenance throughout the development and testing process. In particular, the code generation of rules produced files describing neural network weights, which drastically streamlined the creation process that would have been too complex and error prone manually.

\section{\uppercase{Discussion, Conclusion, and Future Work}}
\label{sec:discussion conclusion future work}

\subsection{\uppercase{Discussion}}

HyperGraphOS marks a major advancement in OSs, tailored for scientific and engineering applications. Leveraging a web-based architecture and DSLs, HyperGraphOS offers a flexible and robust platform for managing complex models and data. This section compares HyperGraphOS to other state-of-the-art systems and highlights its unique contributions. In comparison to systems like PlantUML \cite{correia2024towards} and Graphviz \cite{gansner2009drawing}, which are widely used for static diagram creation and visualization, HyperGraphOS distinguishes itself by enabling dynamic interactions with graph models and seamless integration with advanced technologies like AI and LLMs \cite{camara2023assessment}. The ability to manipulate nodes and links programmatically using JavaScript and navigate virtually unlimited workspaces sets HyperGraphOS apart from traditional graph modeling tools.

When compared to DSL-centric systems like MetaEdit+ \cite{tolvanen2016model}, JetBrains MPS \cite{pech2013jetbrains}, and Eclipse Xtext \cite{herrera2014enhancing}, HyperGraphOS provides a more intuitive and accessible interface due to its web-based architecture and extensive use of the visual OmniSpace. Its flexibility in creating and adapting DSLs enables rapid prototyping and incremental development, which is especially beneficial for dynamic research projects. Furthermore, HyperGraphOS's integration with AI components and LLMs facilitates complex task automation and significantly boosts productivity \cite{sadik2023analysis}. By offering on-demand assistance and intelligent data manipulation within documents, HyperGraphOS expands the possibilities of AI-augmented OmniSpace.

\subsection{\uppercase{Conclusion}}

In this paper, we presented HyperGraphOS, a modern DSL-based OS designed specifically for scientific and engineering interdisciplinary applications. Through the use of DSLs and graph-based model representations, HyperGraphOS provides users with an intuitive and flexible platform for creating, manipulating, and visualizing complex models and data. HyperGraphOS's open-source nature invites further exploration and contributions from the community. The tool is available as open-source software and can be accessed at \cite{hypergraphos-repo}, where additional documentation and videos \cite{hypergraphos-documentation} and future updates are posted.

The case studies in robotic task planning, dynamic research projects, and virtual receptionist systems demonstrate HyperGraphOS’s versatility and practical benefits. In comparing HyperGraphOS to other state-of-the-art systems, its unique contributions were outlined, such as dynamic graph model interaction, flexibility in creating new DSLs, and seamless integration with AI components. While HyperGraphOS presents numerous advantages, it also opens up opportunities for further enhancements, particularly in data handling, scalability, and facilitating collaboration. Expanding these capabilities will be essential as the system evolves to handle increasingly complex and larger datasets.

\subsection{\uppercase{Future Work}}

Moving forward, there are several areas for improvement in HyperGraphOS. While relying on external services for security and privacy management offers flexibility, future iterations could include robust, built-in security measures to strengthen data protection. As the system scales to support larger and more complex datasets, enhancing performance while maintaining a seamless user experience will be essential.

There is also significant potential in further exploring low-code development platforms and innovation. For instance, a start-up like Thunkable \cite{thunkable} provides a no-code platform for designing and creating mobile applications. Its drag-and-drop interface and pre-built components allow users, even without a development background, to create fully functional iOS and Android.
Another innovative example is the Rabbit R1 \cite{rabbit_r1}, which introduces an AI-driven OS designed to simplify interactions with apps and services through voice commands and AI-powered tools, positioning it as a next-generation alternative to smartphones and smart speakers. A drawback of OmniSpace is its reliance on large screens for comfortable use. One potential mitigation could involve enabling the use of HyperGraphOS in Virtual Reality settings. HyperGraphOS shows great potential but requires further development in several areas. For this reason, HyperGraphOS will soon become an open-source project to garner support from early adopters.

In summary, HyperGraphOS presents a novel approach to OS design that addresses the evolving needs of modern scientific and engineering workflows. Its flexible, efficient, and user-friendly platform sets the stage for further advancements in OS design.

\bibliographystyle{splncs04}
{\small
\bibliography{Main}
\end{document}